\documentclass[11pt]{article}

\newif\ifdebug
\debugtrue

\usepackage{cite} 
\usepackage{booktabs}
\usepackage{blindtext}
\usepackage[linesnumbered,lined,ruled,vlined]{algorithm2e}
\SetAlFnt{\scriptsize}

\SetCommentSty{mycommfont}
\usepackage{capt-of}
\usepackage{wrapfig}
\usepackage{subcaption}
\usepackage{epstopdf} 
\usepackage{floatrow} 
\usepackage[percent]{overpic}
\usepackage[table]{xcolor}
\usepackage{tabularx}
\usepackage{setspace}
\usepackage{float}
\usepackage{textcomp}
\usepackage{epsfig}
\usepackage{esvect}
\usepackage{graphicx} 
\usepackage{tikz} 
\usepackage{comment} 
\usepackage{amsmath,amssymb} 
\usepackage{xfrac} 
\usepackage{color} 
\usepackage{bibentry}
\usepackage[inline]{enumitem} 
\usepackage[normalem]{ulem}
\usepackage[symbol]{footmisc}

\nonstopmode

\usepackage[pdfdisplaydoctitle, bookmarksdepth=2]{hyperref}

\newcommand{\xtilde}{{\footnotesize$\sim$}}

\makeatletter\@ifundefined{etal}{\def\etal{\textit{et~al.}}}{}\makeatother

\hypersetup{
pdftitle={Proximally Sensitive Error for Few-Shot Anomaly Detection and Feature Learning},
pdfauthor={Amogh Gudi, Fritjof B\"{u}ttner, Jan van Gemert}
}

\begin{document}

\date{} 

\title{Proximally Sensitive Error\\for Anomaly Detection and Feature Learning}

\author{Amogh Gudi, Fritjof B\"{u}ttner, Jan van Gemert}

\maketitle
    \begin{abstract}
\label{sec:abstract}
Mean squared error (MSE) is one of the most widely used metrics to expression differences between multi-dimensional entities, including images.
However, MSE is not locally sensitive as it does not take into account the spatial arrangement of the (pixel) differences, which matters for structured data types like images.
Such spatial arrangements carry information about the source of the differences; therefore, an error function that also incorporates the location of errors can lead to a more meaningful distance measure.
We introduce Proximally Sensitive Error (PSE), through which we suggest that a regional emphasis in the error measure can `highlight' semantic differences between images over syntactic/random deviations.
We demonstrate that this emphasis can be leveraged upon for the task of anomaly/occlusion detection. 
We further explore its utility as a loss function to help a model focus on learning representations of semantic objects instead of minimizing syntactic reconstruction noise.

\end{abstract}

	\section{Introduction}
\label{sec:intro}

\begin{figure}[t]
	\centering
	\centerline{\includegraphics[width={0.75\linewidth}]
		{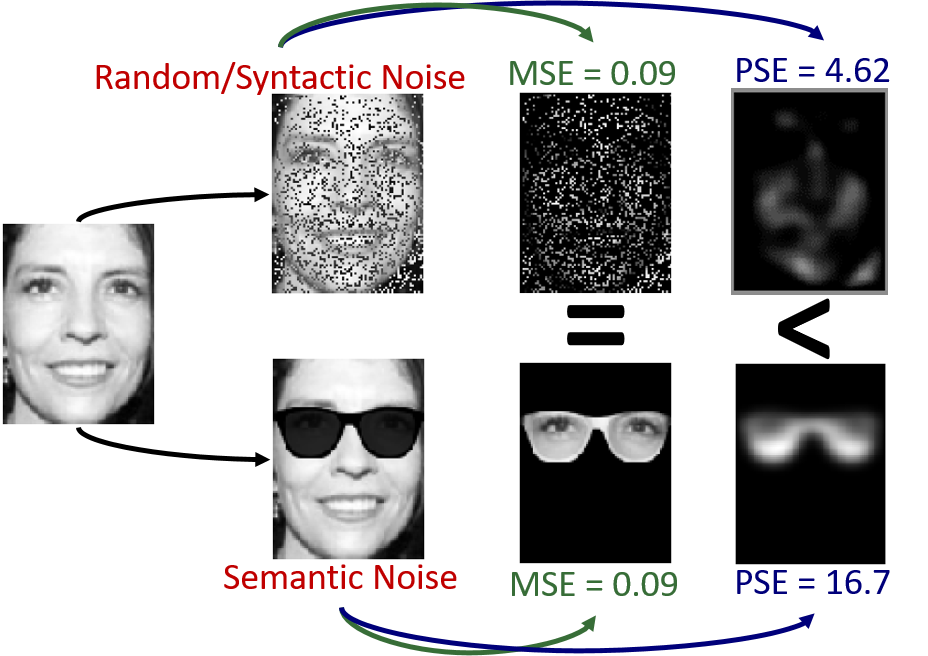}}
	\caption{Comparison PSE vs MSE: A face image (left) is subtracted by two variants (col 2): one with sunglasses (bottom), and one with some random noise addition (top). MSE between the original image and the two variants is equal (col 3). However, PSE of the image with sunglasses is much higher than with random noise (col 4). This illustrates that MSE cannot distinguish between syntactic/random and semantic/meaningful errors, while PSE can.}
	\label{fig:PvMrecon}
\end{figure}

Since its introduction by Eukleídēs~\cite{math} in \xtilde300BC, Euclidean/L2 distance has widely been used to express distances between points. 
The simplicity, robustness, and mathematical convenience of this metric makes it a popular choice as a similarity/distance metric in computer vision and machine learning in the form of mean squared error (MSE)~\cite{lee2013study,ng2011sparse,kingma2013auto}, where it finds wide use for all domains of data types including images~\cite{mao2016multi,chen2017deep}.
In this work, we present an alternate to MSE for structured data types (images) in the form of a proximally sensitive error function. 

MSE is well suited for measuring differences between scalars which are single-dimensional, or vectors which are multi-dimensional.
Such data types have no information in the spatial arrangement of values/elements within, i.e., there are no patterns in the way their elements are located (they are only required to be consistent). For example, a feature vector describing an object's important characteristics. 
For such non-structured data types, the relative locations (indexes) of the differences does not matter. 
On the other hand, for structured data types like images, relative locations of the differences do matter because information is also encoded in the the underlying spatial arrangement of values.
For example, pixels composing a visual image of an object. 
However, this is not considered by the mean squared error, as illustrated with an example in Figure~\ref{fig:PvMrecon}~(column 3).
In this study, we attempt to address this drawback of MSE by introducing a locally sensitive metric.



An error function that incorporates spatial location of errors can lead to a more meaningful error metric for images, since the meaning of the content of an image relies heavily on the location of the pixel values that represent semantic or meaningful objects. 
Based on the observation that adjacent pixels often form regions of semantic meaning (in other words: objects in the image) while sparse spread-out errors are caused by random or syntactic noise, we hypothesise that an error function that emphasizes regions with high concentrations of pixel-wise reconstruction error forms a better metric (see example in Figure~\ref{fig:PvMrecon}, column 4).
Towards this end, we introduce Proximally Squared Error (PSE), that implements this spatial dependence via Gaussian convolutions. 

In the field of computer vision, one of the utilities of such a location sensitive error metric can be in tasks involving image reconstruction.
Typically for such tasks, the difference between a reconstructed image and the original image is computed to iteratively improve the reconstruction~~\cite{mao2016multi,zhou2017anomaly} and/or make a downstream classification~\cite{wiehman2016unsupervised,caron2019unsupervised}. 
To empirically evaluate our hypothesis, we examine proximally sensitive error against mean squared error for the tasks of image anomaly detection and unsupervised pre-training, both involving image reconstruction.

\paragraph{Contributions}
This work has the following contributions:
\begin{enumerate}[label=(\roman*)]
	\item We present proximally sensitive error (PSE), a novel locally sensitive error function that also takes into account the relative locations of differences between such structured data types.
	\item We examine and provide insights into the applicability of PSE versus MSE for image reconstruction powered tasks of anomaly detection and unsupervised pre-training.
\end{enumerate}

	\section{Related Work}
\label{sec:relwork}
\paragraph{Distance metrics for images}
The straight-forward pixel-wise L2/Euclidian distance, equivalent to the mean squared error (MSE), is a popular choice for expressing differences between images \cite{ng2011sparse,kingma2013auto, chen2017deep, mao2016multi}.
MSE has been used in the image reconstruction terms of loss functions for training several neural network approaches such as sparse autoencoders~\cite{ng2011sparse}, variational autoencoders~\cite{kingma2013auto}, convolutional autoencoders~\cite{chen2017deep}, and generative adversarial networks~\cite{mao2016multi}. 
In this work, we propose proximally sensitive error (PSE) as an alternative to MSE to express differences between images.

MSE does not match well with visually perceived differences between images by humans~\cite{girod1991psychovisual, sinha2011perceptually}.
To counter this, Wang \etal~\cite{wang2004image} introduced the structured similarity metric (SSIM), an distance metric specifically designed for assessment of image/video quality loss due to compression. SSIM is able to look at neighbouring pixels within a pre-defined window.
Concurrently, Li \etal~\cite{li2003discovery} discovered heuristics from a large dataset to design the dynamic partial distance function (DPF) that better represents perceptual similarity. 
In our work, we propose a simplified distance function for structured data like images by incorporating regional sensitivity. 

\paragraph{Image anomaly detection}
Anomaly detection in images can be divided into two broad classes (among others) \cite{pang2021deep,ehret2019image}: direct inference/classification of anomalies \cite{liznerski2020explainable, venkataramanan2020attention, golan2018deep}; and reconstruction and comparison based anomaly detection (typically un/semi-supervised) \cite{hawkins2002outlier, chen2017outlier, zhou2017anomaly}. 
Bergmann \etal~\cite{bergmann2021mvtec} examined SSIM as a distance metric for autoencoder reconstruction anomaly detection, which yielded results similar to L2.
In our work, we focus on a PCA reconstruction based anomaly detection setup where we examine the applicability of PSE as a distance metric.

\paragraph{Unsupervised pre-training}
Unsupervised pre-training was discovered to provide a superior alternative to fully stochastic parameter initialization of deep neural networks \cite{bengio2007greedy, erhan2009difficulty, erhan2010does}. 
The use of end-to-end pre-training on image data has been widely studied due to its potential benefits \cite{wiehman2016unsupervised, caron2019unsupervised}, typically using the pixel-wise L2/MSE reconstruction loss for optimization. 
In this paper, we explore PSE as a location-sensitive image reconstruction loss for unsupervised pre-training with images, and examine if this can lead to better feature learning.

	\section{Method}
\label{sec:meth}

\begin{figure}
	\centering
	\centerline{\includegraphics[width={\linewidth}]
		{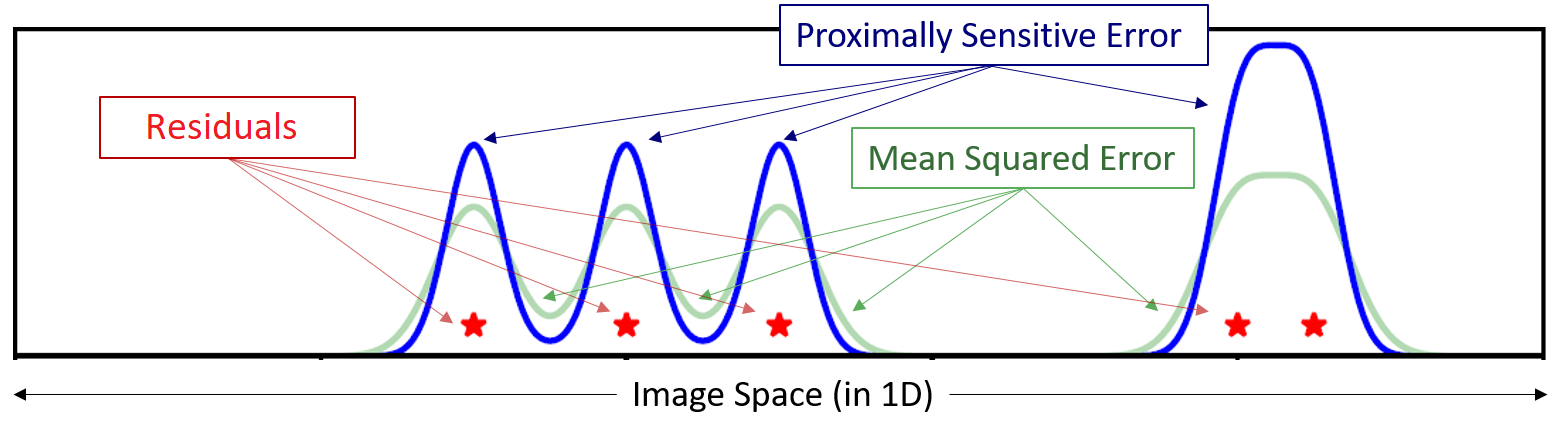}}
	\caption{Illustration MSE vs PSE: MSE magnitude in the areas with more spread-out errors is the same as areas with concentrated grouping of residuals. In contract, PSE error is higher for closely grouped areas.}
	\label{fig:PvM1D}
\end{figure}

For two-dimensional images, the residuals between two images is simply the differences between individual pixel values of the two images. This can be expressed mathematically as:
\begin{equation}
\mathcal{R}_{i,j} = \hat{Y}_{i,j} - Y_{i,j} \qquad\forall~~i \in M,~~j \in N
\end{equation}
where $Y_{i,j}$, $\hat{Y}_{i,j}$ represents the $i$\textsuperscript{th} and $j$\textsuperscript{th} pixel in images $Y$ and $\hat{Y}$ of size $M \times N$.

The Mean Squared Error (MSE) between these two images is essentially the average of the squares of the residuals between the two images. This is expressed as:
\begin{equation}
\text{MSE} = \frac{1}{MN} \sum_{i,j=1}^{M,N} \left(\mathcal{R}_{i,j}\right)^2
\end{equation}

As can be seen, the squared error values do not take the neighbourhood of individual errors into consideration. Thus, residuals that are grouped tightly and residuals that are spread-out contribute equally towards the mean. This is illustrated in Figure \ref{fig:PvM1D}.

To counter this, we propose a Proximally Sensitive Error (PSE) function:
\begin{equation}
\text{PSE} = \frac{1}{MN} \sum_{i,j=1}^{M,N} \left( \left[ \mathcal{R}*k(\sigma)\right]_{i,j}\right)^2
\end{equation}
where $*$ denotes the convolution operation and $k(\sigma)$ denotes a Gaussian kernel with $\sigma$ as its standard deviation, given by:
\begin{equation}
k(\sigma)_{x,y} = \frac{1}{2\pi\sigma^2}e^{-\frac{x^2+y^2}{2\sigma^2}} \qquad\forall~~x,y \in 2\sigma
\end{equation}

Through the convolution operation, every error value incorporates the error value in its local neighbourhood (defined by $\sigma$). This way, PSE is able to give a higher importance to residuals grouped close together versus residuals sparsely spread-out. This is also visualized in Figure \ref{fig:PvM1D}.

\subsubsection{Image reconstruction based anomaly detection}
\begin{figure}
	\centering
	\centerline{\includegraphics[trim={0 10.4cm 2.6cm 0},clip,width={\linewidth}]
		{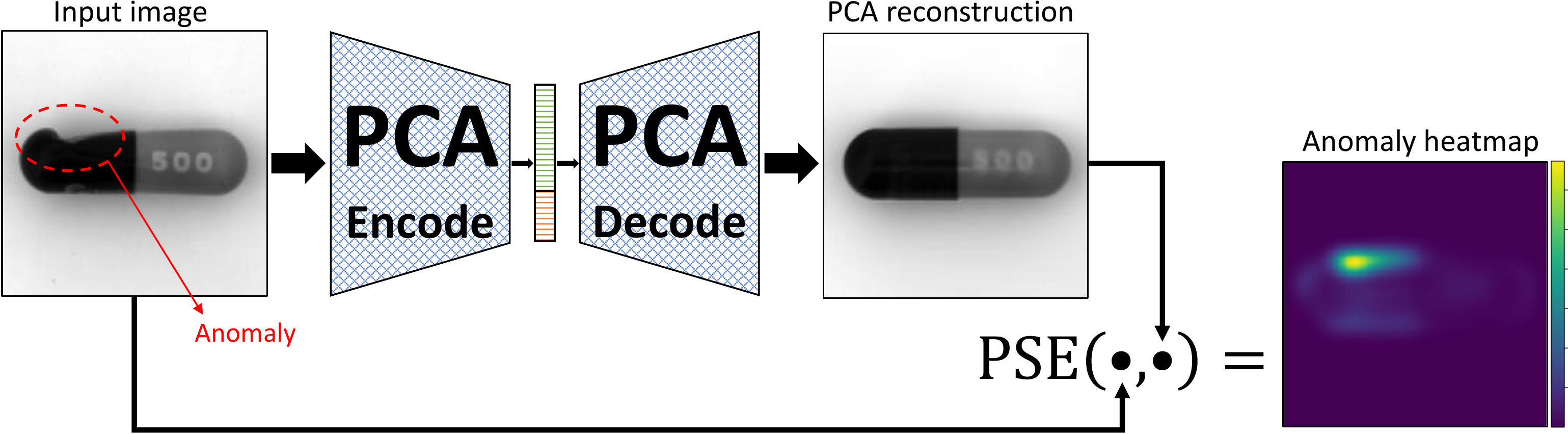}}
	\caption{Illustration of the anomaly detection pipeline using proximally sensitive error (PSE). A PCA model trained on non-anomalous images is unable to reconstruct anomalies present in the input (the crushed part of the capsule in this example). PSE can be used to highlight this difference, as seen in the produced heatmap.}
	\label{fig:illusano}
\end{figure}

Assuming closely grouped errors denote the more meaningful/semantic differences between images (our hypothesis), we can leverage upon PSE's ability to highlight them to perform anomaly detection.
The pipeline for anomaly detection based on image reconstruction involves the following steps, and illustrated in Figure \ref{fig:illusano}.
\begin{itemize}
	\item A principle component analysis (PCA) model \cite{hotelling1933analysis}, which is capable of image reconstruction, is trained on purely non-anomalous images of a particular object/class.
	
	\item The input image is passed through this PCA model and its reconstructed image is obtained.
	Due to the model's training, the non-anomalous parts of the input image are reconstructed correctly, but the model fails to reconstruct any anomalies in the image that it hasn't seen during its training. 
	This effect can further be enhanced by reducing the number of PCA components used to reconstruct the image.

	\item Next, the pixelwise PSE (with a set $\sigma$ parameter) between the original and reconstructed image is computed. This pixelwise PSE `image' can essentially be seen as a heatmap of differences between the two images. 
	
	\item Finally, the maximum PSE value in this computed heatmap is chosen denoting the area with the highest difference, and this value provides and estimate of anomalousness in the image.
\end{itemize}
The $\sigma$ parameter of the PSE function and the number of PCA components used for image reconstruction can be optimized/learnt (e.g., via grid search) based on examples of anomalous and non-anomalous images available in the training set.

\subsubsection{Feature learning via unsupervised pre-training}
If the closely-grouped errors assigned a higher value by PSE are indeed more semantic in nature (our hypothesis), PSE can essentially perform as a more meaningful loss function for unsupervised image reconstructing. 
This is because the model can be made to focus more on minimizing the reconstruction of the closely grouped semantic errors over the spread-out syntactic errors during training, leading to better feature learning.

Such unsupervised image reconstruct can be used as a pre-training step for a downstream task such as classification using a autoencoder-style neural network. The steps involved in this setup are as follows given a semi-supervised dataset of unlabelled and labelled samples:
\begin{itemize}
	\item First, an autoencoder is trained end-to-end for image reconstruction with PSE as the loss function using the large set of unlabelled samples. 
	\item Next, the encoder part of the autoencoder is appended with classification layer(s) (e.g. fully connected layer with softmax activation) while the decoder is discarded. 
	This new model is now trained on the smaller set of labelled samples. 
\end{itemize}
Image classification can now be performed with this trained encoder-classifier model.

	\section{Experiments and Results}
\label{sec:exp}

\subsection{Anomaly/Occlusion Detection}
We focus on the task of few-shot anomaly detection in these experiments. 

\paragraph{Setup}
The task of anomaly detection consists of classifying if a particular object in a given input image has an abnormality. 
For example, picture of a bottle with a crack in it can be classified as anomalous since normal bottles do not have cracks. 
Under the few-shot learning regime for anomaly detection, the vast majority of available training samples belong to the non-anomalous class, and only a handful of samples contain anomalies.

\paragraph{Datasets}
We perform this set of experiments on two publicly available datasets: The MVTecAD dataset \cite{bergmann2021mvtec} for industrial anomaly detection, and the AR Face dataset~\cite{ar} for facial occlusion detection.
The MVTecAD dataset consists of \xtilde5300 images of 15 categories of object classes, with \xtilde4100 non-anomalous samples and \xtilde1200 images with multiple types of anomalies. 
The AR Face dataset consists of 2600 frontal images of faces with varying illumination and facial expressions, with 1200 of them containing occlusions caused by sunglasses and scarf.
Images from both datasets are resized and converted to 128$\times$128 grayscale.
Examples from these datasets can be seen in Figure~\ref{fig:ex-ad} (first rows).

\subsubsection{Industrial Anomaly Detection}
The results of 5-shot industrial anomaly detection on the MVTecAD dataset are shown in Figure \ref{fig:res-iad} in terms of average precision. 
Only 5 labelled samples per anomaly class are used for optimizing the $\sigma$ hyper-parameter of the PSE function and the number of PCA components used for image reconstruction. 

As can be seen, the use of PSE attains a higher average precision (0.73$\pm$0.2) than MSE (0.66$\pm$0.3) on average.
Better performance is obtained by PSE w.r.t MSE on 11 of the 15 object categories in the dataset.
In few categories like \textit{toothbrush} and \textit{bottle}, the performance gap between PSE and MSE is marginal while the performance is very high. 
This is likely due to the straight-forward appearance of anomalies in the images, thereby making the task trivial for both PSE and MSE.

\begin{figure}[t]
	{\subfloat[][MVTecAD Industrial Dataset]{\includegraphics[width=\textwidth]{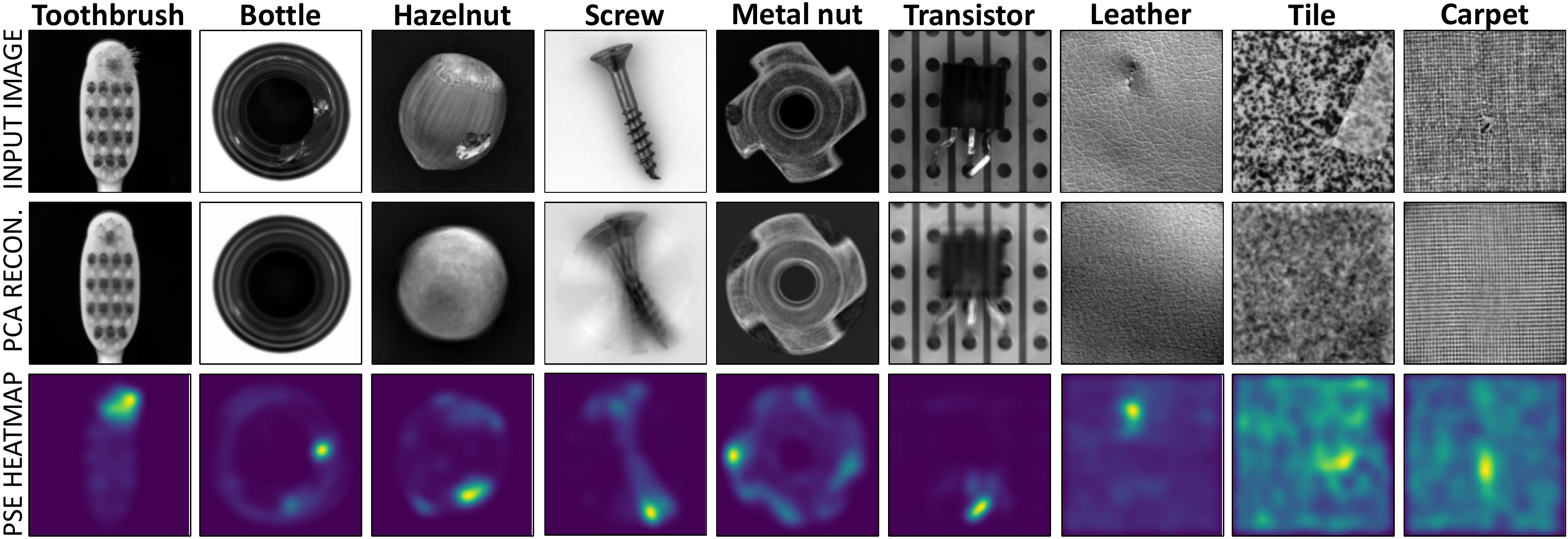}\label{fig:ex-iad}}\vspace{0.1cm}} %
	\floatbox[{\capbeside\thisfloatsetup{capbesideposition={right,bottom},capbesidewidth=0.52\textwidth}}]{figure}[\FBwidth]
	{\subfloat[][AR Face Dataset]{\includegraphics[width=0.45\textwidth]{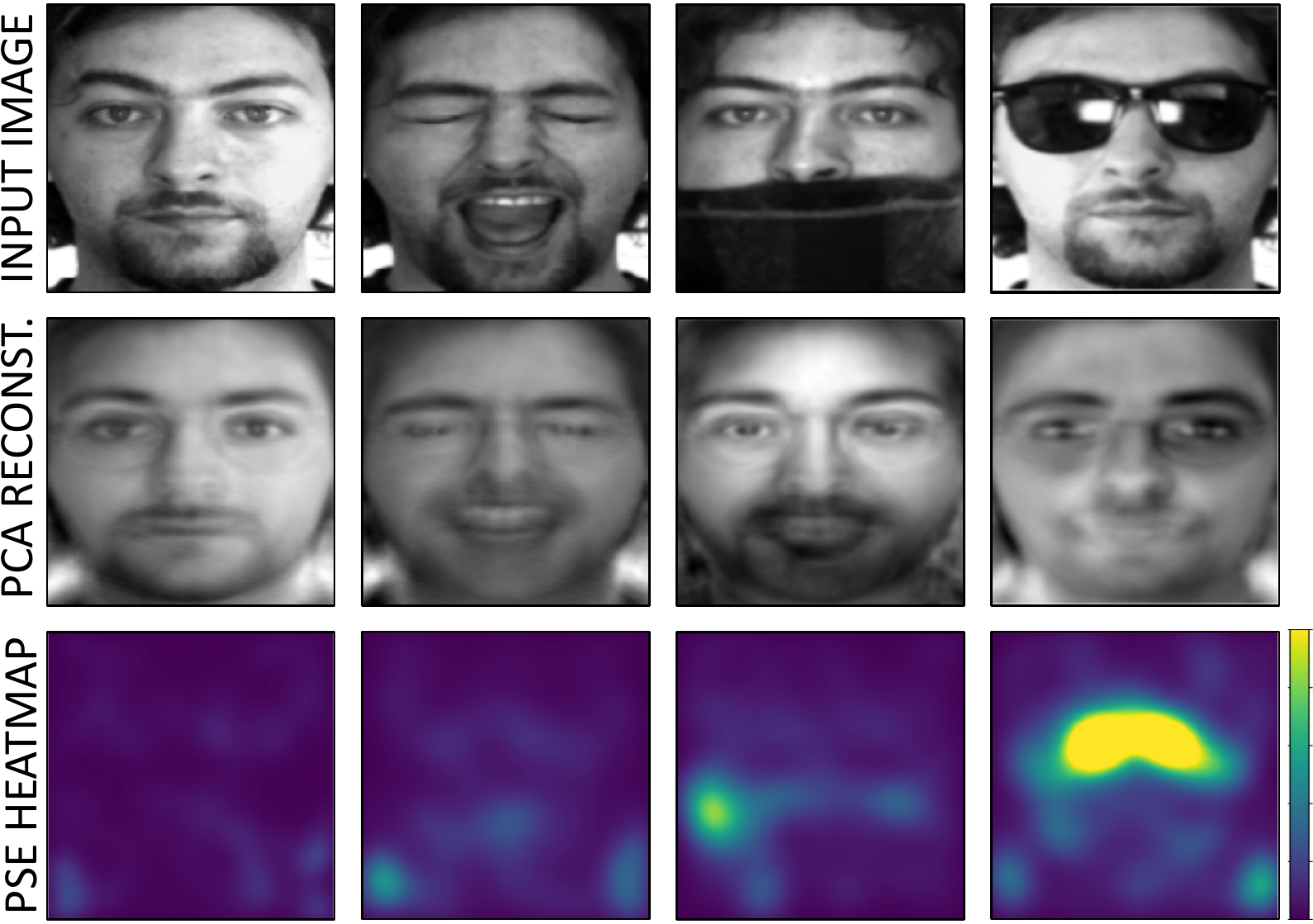}\label{fig:ex-fod}}}
	{\caption{
		Examples of 5-shot anomaly/occlusion detection on the MVTecAD industrial dataset (\subref{fig:ex-iad}) and AR face dataset (\subref{fig:ex-fod}).
        The PCA is unable to reconstruct (middle rows) the anomaly in the input image (top rows). Using this, PSE essentially provides a heatmap of the anomaly (bottom rows).
        Reconstructions of \textit{screw} and \textit{metal nut} are especially poor due lack of pose-normalization in the images. 
		} 
	\label{fig:ex-ad}}
\end{figure}

On the other hand, images from two of the categories where PSE performer poorer than MSE (\textit{hazelnut} and \textit{screw}) are not pose-normalized and contain rotation. 
This results in a very high variation in terms of the object's pose, which a simple PCA model is unable to model and reconstruct. 
For images of the \textit{tile} category, some anomalies are in the form of transparent occlusions which leads to unreliable performance. 
Lastly, anomaly detection performance for \textit{cable} is the lowest for both MSE and PSE. 
This is because most anomalies in this category are in the colour space (e.g., differently coloured wires are swapped), and this information is lost due to the grayscale conversion.
These observations can be seen in the examples shown in Figure \ref{fig:ex-ad}.


\subsubsection{Facial Occlusion Detection}
\begin{figure}
	{\subfloat[][MVTecAD Industrial Dataset]{\includegraphics[width=\textwidth]{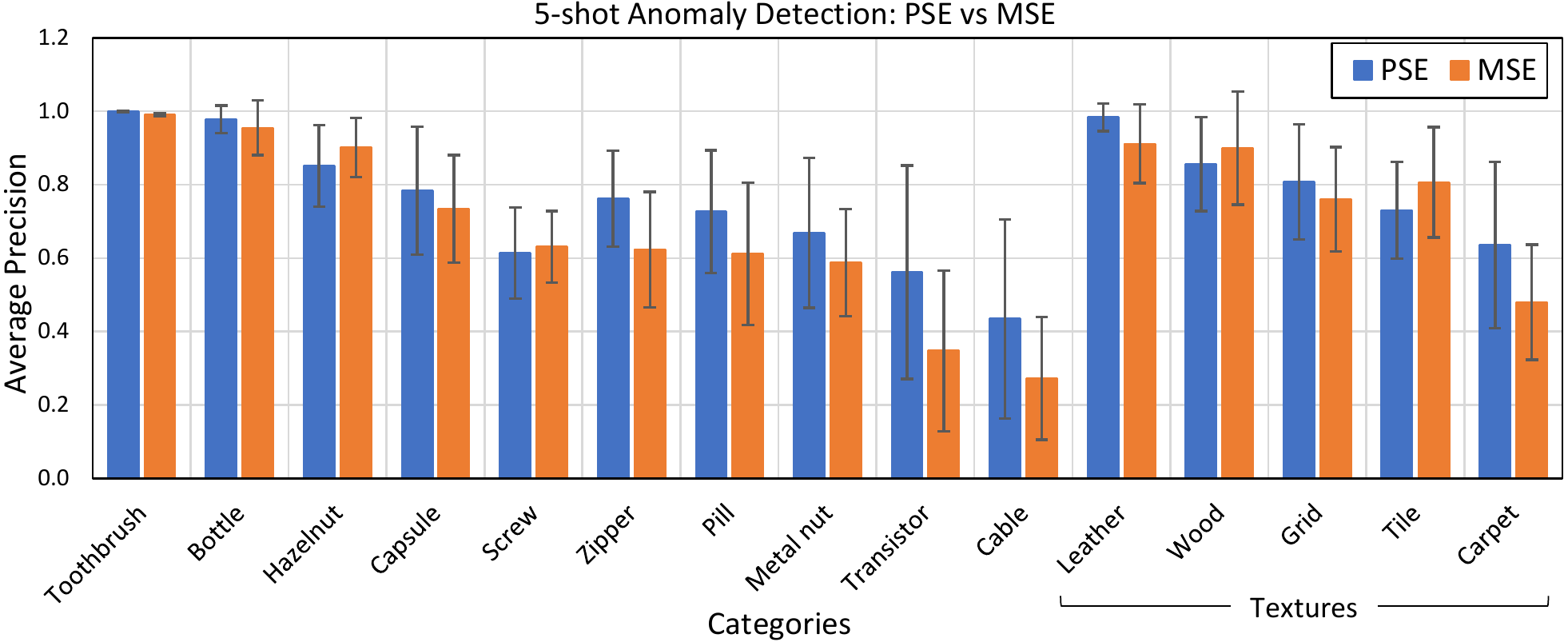}\label{fig:res-iad}}} 
	\floatbox[{\capbeside\thisfloatsetup{capbesideposition={right,bottom},capbesidewidth=0.5\textwidth}}]{figure}[\FBwidth]
	{\subfloat[][AR Face Dataset]{\includegraphics[width=0.36\textwidth]{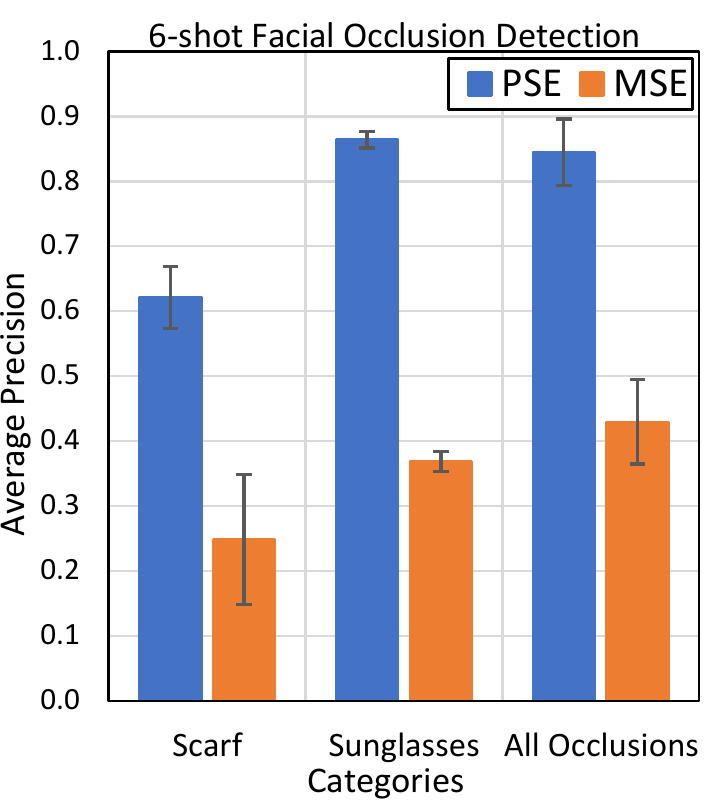}\label{fig:res-fod}}}
	{\caption{
		Results of 5-shot anomaly/occlusion detection on the MVTecAD industrial dataset (\subref{fig:res-iad}) and AR face dataset (\subref{fig:res-fod}).
        PSE performs better than MSE in almost all categories. 
        Poor performance in some categories are due lack of pose normalization (\textit{screw}, \textit{hazelnut}), transparent occlusions (\textit{tile}), loss of colour information (\textit{cable}), and higher variation in lower face (\textit{scarf}).
        } 
	\label{fig:res-ad}}
\end{figure}

Figure \ref{fig:res-fod} summarises and compares the results of 5-shot facial occlusion detection on the AR Face dataset using PSE and MSE. Similar to the previous experiment, 5 labelled samples per class are used to determine the model hyper-parameters ($\sigma$ and PCA components).

It can be seen that PSE significantly outperforms MSE: PSE obtains an average precision of $0.84\pm0.05$ while MSE scores $0.43\pm0.07$ over all classes.
Both PSE and MSE perform better with detecting sunglasses as compared to scarf, likely due to the scarf occluding the lower part of the face that contains more variation due to facial hair and expressions. 

Overall, PSE based occlusion detection on faces from the AR dataset works a lot better than anomaly detection on the MVTecAD dataset. 
This can be attributed to the pose-normalisation of face crops in AR and the much higher number of training samples available, as compared to that of the individual object categories of MVTecAD.

\subsection{Unsupervised Pre-training}
In these experiments, we evaluate the use of PSE vs MSE as a loss function for the task of unsupervised pre-training. 

\begin{figure}
	\centering
	\centerline{\includegraphics[width={0.85\linewidth}]{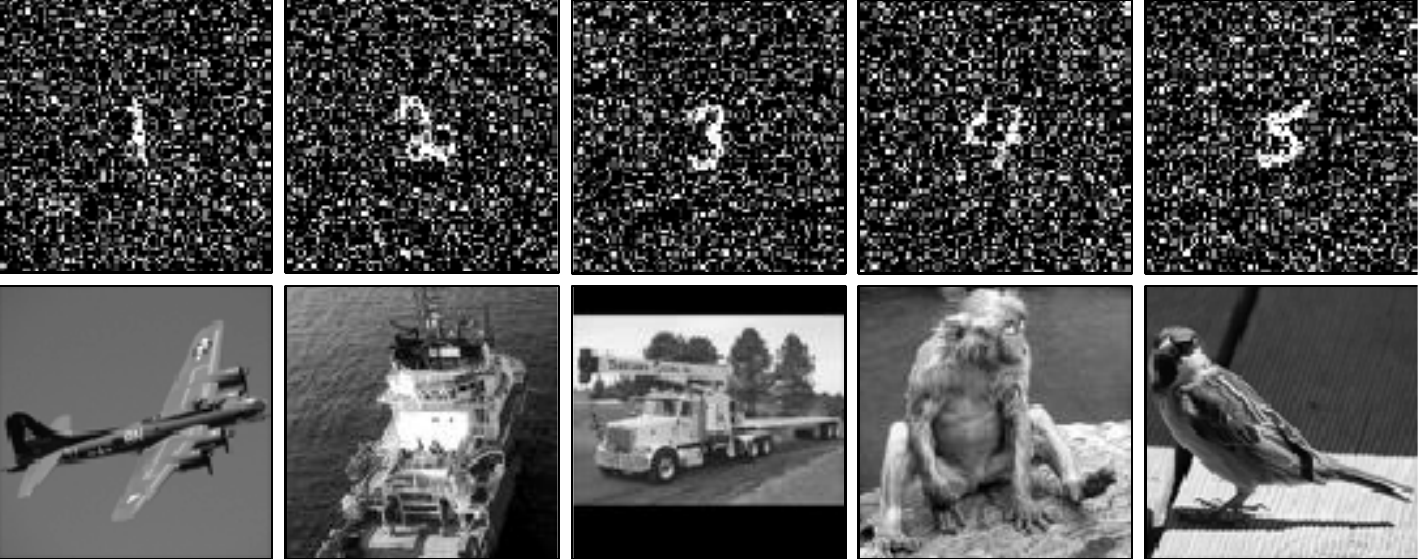}}
	\caption{Example images from the modified MNISTX (top row) and STL-10 datasets (bottom row). The depicted classes are (left to right): 1, 2, 3, 4, 5 from MNIST; and airplane, ship, truck, monkey, bird from STL-10.}
	\label{fig:ex-upt}
\end{figure}

\paragraph{Datasets}
Two classification datasets are used in this set of experiments: a modified version of MNIST handwritten digits dataset \cite{lecun1998gradient} (hereby called MNISTX) composed of 70,000 labelled samples (10 classes); and a grayscale version of the STL-10 image classification dataset \cite{stl10} containing 100,000 unlabelled and 500 labelled images (10 classes).
Examples from both datasets are shown in Figure~\ref{fig:ex-upt}.
The MNIST dataset has been modified by padding the original image such that its size is tripled (84$\times$84 pixels) and adding salt-and-pepper noise to the image.
The STL-10 dataset images are 96$\times$96 pixels and converted to grayscale. 

\paragraph{Setup}
We consider a simple autoencoder neural network architecture composed on an input layer, a fully connected hidden layer with ReLU that serves as the bottleneck, and a fully connected output layer with a sigmoid activation function.
Another fully-connected softmax layer serves as the classification layer whose dimensions match the number of classes in the dataset, i.e., 10. 
The dimensions of the input and output layers match the number of pixels in the input images (i.e., 84$\times$84 for MNISTX and 96$\times$96 for STL-10).
The $\sigma$ parameter of the PSE loss function is set to 0.5 (determined empirically). 

\subsubsection{Computational Efficiency} 
\begin{figure}
	\centering
    \subfloat[][MNISTX]{\includegraphics[width=.45\textwidth]{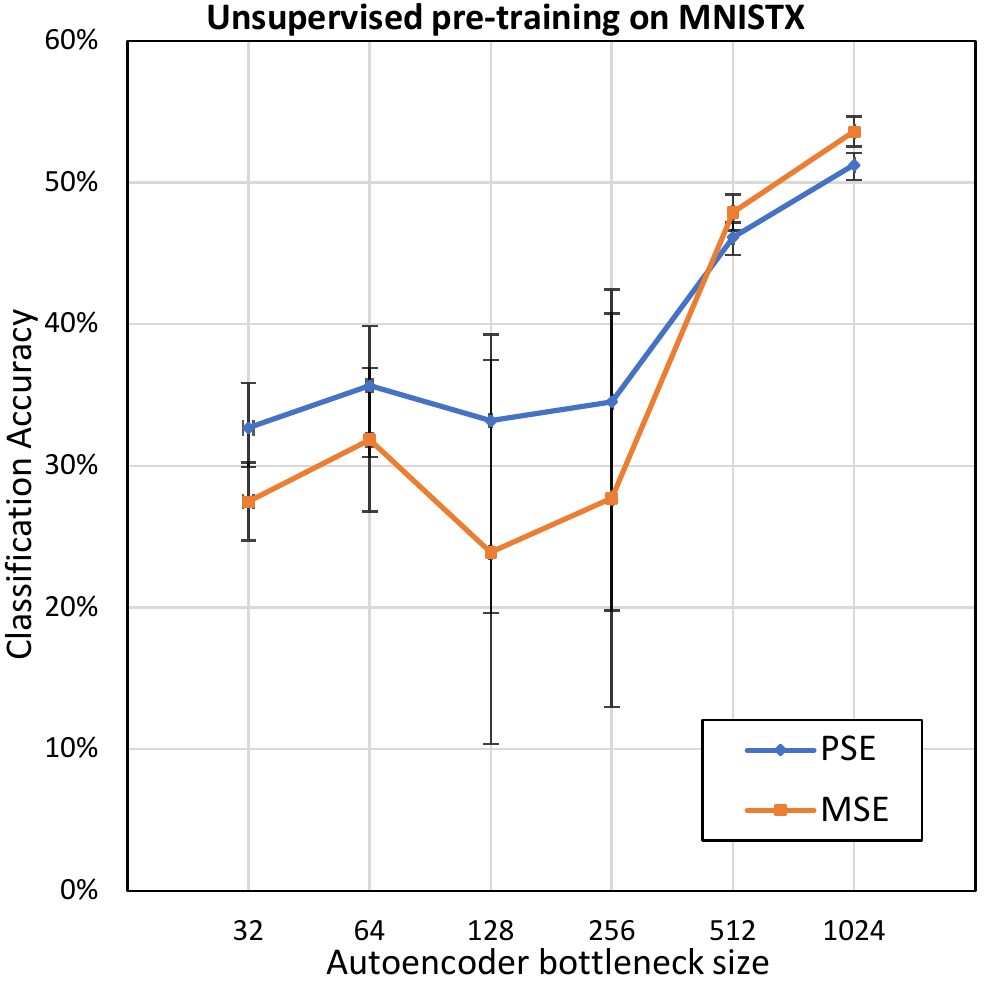}\label{fig:res-mnist-ce}}\qquad
	\subfloat[][STL-10]{\includegraphics[width=.45\textwidth]{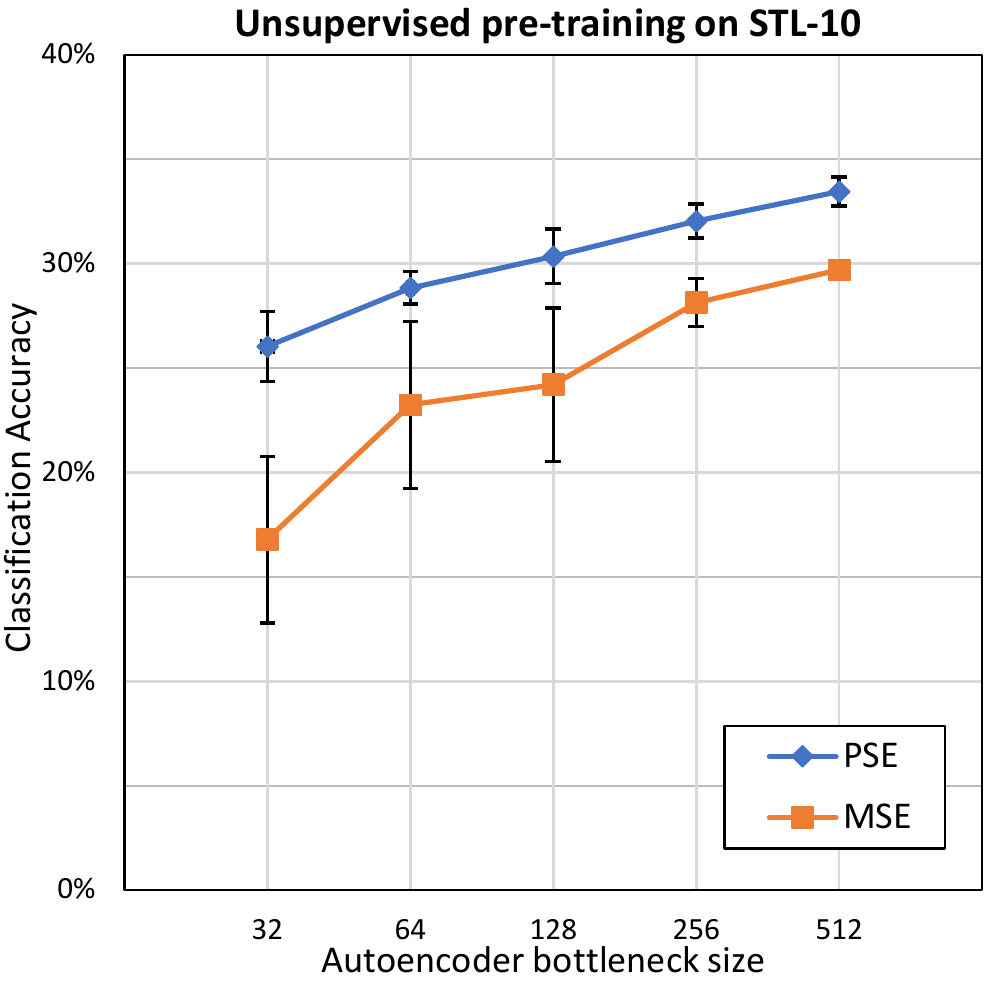}\label{fig:res-stl10-ce}}
	\caption{Results of unsupervised pre-training using PSE and MSE on MNISTX (\subref{fig:res-mnist-ce}) and STL-10 (\subref{fig:res-stl10-ce}) datasets in terms of the autoencoder bottleneck size (error bars represent standard deviations). 
    PSE performs better than MSE for models with smaller bottlenecks, but this gap reduces for higher complexity models.}
	\label{fig:res-ce}
\end{figure}

The resulting accuracy obtained by the model using PSE and MSE with respect to the autoencoder bottleneck size (i.e., dimension of the hidden layer) can be seen in Figure \ref{fig:res-ce}.
For both datasets, it can be seen that using PSE results in higher accuracy than MSE on average per bottleneck size.
This means PSE can achieve the same results as MSE in spite of using smaller models.
In the STL-10 dataset, the accuracy gap between PSE and MSE is largest when the autoencoder's bottleneck is the smallest, and this gap reduces gradually as the bottleneck size increases. 
On the MNISTX dataset, such an observation is not as clear (standard deviations overlap), however the accuracy gap does become smaller/inverts for large bottleneck sizes. 

These observations can be attributed to the the fact that a model with fewer parameters benefits the most from PSE's ability to focus the learning on minimizing semantic errors, thereby leading to more computationally efficient feature learning. 
On the other hand, the higher learning capacity of larger models might enable them to learn meaningful features in spite of no explicit boosting of semantic errors over syntactic ones in MSE.

\subsubsection{Data Efficiency} 
\begin{figure}
	\centering
    \subfloat[][MNISTX]{\includegraphics[width=.45\textwidth]{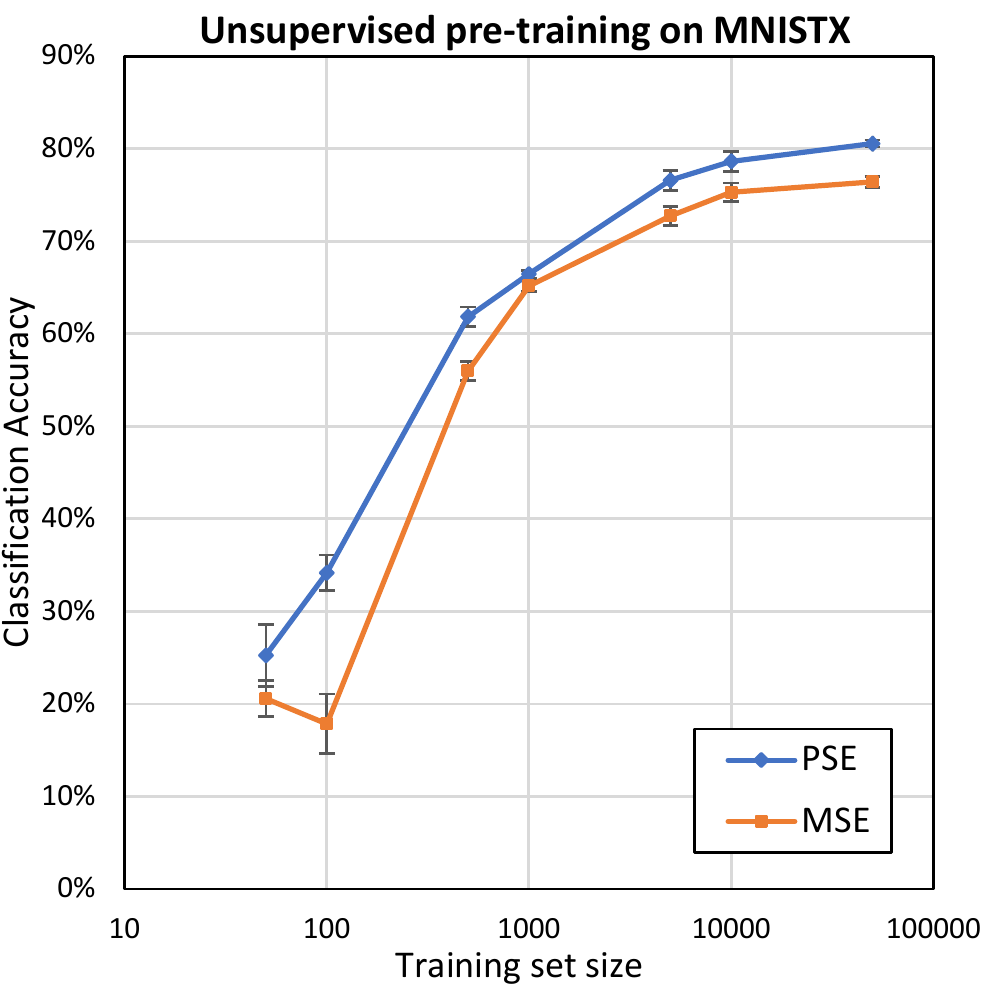}\label{fig:res-mnist-de}}\qquad
	\subfloat[][STL-10]{\includegraphics[width=.45\textwidth]{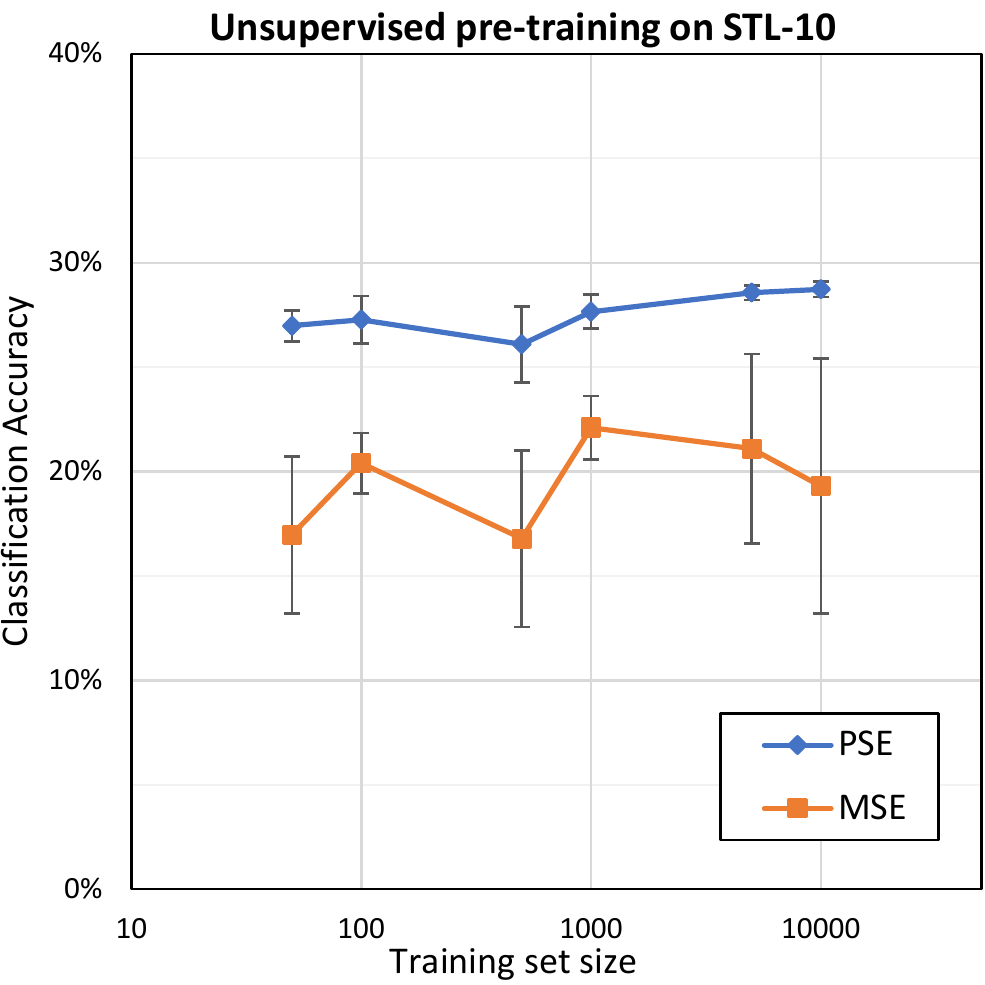}\label{fig:res-stl10-de}}
	\caption{Results of unsupervised pre-training using PSE and MSE on MNISTX and STL-10 datasets in terms of the size of the pre-training set (error bars represent training sets). 
    Overall, PSE performs better than MSE over all sizes of training data used. 
    PSE results are also more consistent, evident by the smaller standard deviation error bars.}
	\label{fig:res-de}
\end{figure}

Figure~\ref{fig:res-de} shows the accuracy obtained by using PSE and MSE for varying amounts of unsupervised pre-training data used. 
As can be seen, PSE consistently outperforms MSE as a loss function on both datasets over all sizes of the unsupervised pre-training set.
This suggests fewer data is required by PSE to achieve the same results as MSE.

On MNISTX, the accuracy gap between PSE and MSE is larger for lower training set sizes, and this gap appears to close in when more training data is introduced. 
This could suggest that PSE's focus on more meaningful errors helps the model learn more efficiently from the data, and this effect is magnified when the available training data is limited.

On the STL-10 dataset, such an observation is not apparent. However, PSE seems to produce more consistent results than MSE, both across and within different training set sizes (the standard deviation of accuracies is smaller). 
This could potentially be caused by the strong attention on the more semantic errors in PSE overcoming the effect of stochasticity in the model's parameter initialization that could have lead to learning distractions.

	\section{Discussion}
\label{sec:disc}
Experiments performed on the application of proximally sensitive error (PSE) for few-shot anomaly detection appeared to show promising results. 
PSE was able to outperform mean squared error (MSE) in detecting anomalies on a large majority of object categories. 
This potentially supports our suggestion that closely grouped differences are primarily semantic in nature and PSE is able to boost them in comparison with spread-out differences, which are largely syntactic, i.e., caused by random noise.

In these experiments, the parameter defining the amount of spread of errors (i.e., the $\sigma$ of the Gaussian kernel) did require to be estimated/optimized per object category. 
This suggests that for different types of objects, the spatial definition of semantic objects vs syntactic noise is different. 
Also, inferring/learning this parameter automatically (along with the number of PCA components used) can improve the applicability of the proposed anomaly detection pipeline from few-shot towards zero-shot detection.

Experiments on the use of PSE as a loss function for unsupervised pre-training suggested that PSE can lead to better feature learning than MSE under constrained conditions: when the model's computational capacity is limited, or when the availability of training data is low.
This can make PSE a promising choice for training models for low-powered computational devices and novel image tasks.
However, the results lack certainty and a wider range of experiments on different types of datasets and mode complex model architectures can shed further light on the generalizability of these results.


\bibliographystyle{unsrt}
\bibliography{../references}

\end{document}